\documentclass[conference]{IEEEtran}
\IEEEoverridecommandlockouts
\usepackage{cite}
\usepackage{amsmath,amssymb,amsfonts}
\usepackage{textcomp}
\usepackage[dvipsnames]{xcolor}

\usepackage{times}
\usepackage{soul}
\usepackage{url}
\usepackage[hidelinks]{hyperref}

\usepackage[small]{caption}
\usepackage{graphicx}
\usepackage{booktabs}
\usepackage{algorithm}

\usepackage{array} 
\usepackage{multirow}

\usepackage{moreverb}
\usepackage{rotating}
\usepackage{fancybox}
\usepackage{picinpar,varioref,floatflt}
\usepackage{bm}
\usepackage{textcomp}
\usepackage{float}
\usepackage{url}
\usepackage{tabularx}
\usepackage{balance}
\usepackage[noend]{algpseudocode}

\def\BibTeX{{\rm B\kern-.05em{\sc i\kern-.025em b}\kern-.08em
    T\kern-.1667em\lower.7ex\hbox{E}\kern-.125emX}}
\begin{document}

\title{FedEP: Tailoring Attention to Heterogeneous Data Distribution with Entropy Pooling for DFL
}

\author{
    \IEEEauthorblockN{Chao Feng\IEEEauthorrefmark{1}, Hongjie Guan\IEEEauthorrefmark{1}, Alberto Huertas Celdrán\IEEEauthorrefmark{1}, Jan von der Assen\IEEEauthorrefmark{1}, G\'er\^ome Bovet\IEEEauthorrefmark{2}, Burkhard Stiller\IEEEauthorrefmark{1}}
    
    \IEEEauthorblockA{\IEEEauthorrefmark{1}Communication Systems Group, Department of Informatics, University of Zurich UZH, CH--8050 Zürich, Switzerland \\{[cfeng, huertas, vonderassen, stiller]}@ifi.uzh.ch, hongjie.guan@uzh.ch}
    \IEEEauthorblockA{\IEEEauthorrefmark{2}Cyber-Defence Campus, armasuisse Science \& Technology, CH--3602 Thun, Switzerland gerome.bovet@armasuisse.ch}
}


\maketitle

\begin{abstract}
Non-Independent and Identically Distributed (non-IID) data in Federated Learning (FL) causes client drift issues, leading to slower convergence and reduced model performance. While existing approaches mitigate this issue in Centralized FL (CFL) using a central server, Decentralized FL (DFL) remains underexplored. In DFL, the absence of a central entity results in nodes accessing a global view of the federation, further intensifying the challenges of non-IID data. Drawing on the entropy pooling algorithm employed in financial contexts to synthesize diverse investment opinions, this work proposes the Federated Entropy Pooling (FedEP) algorithm to mitigate the non-IID challenge in DFL. FedEP leverages Gaussian Mixture Models (GMM) to fit local data distributions, sharing statistical parameters among neighboring nodes to estimate the global distribution. Aggregation weights are determined using the entropy pooling approach between local and global distributions. By sharing only synthetic distribution information, FedEP preserves data privacy while minimizing communication overhead. Experimental results demonstrate that FedEP achieves faster convergence and outperforms state-of-the-art methods in various non-IID settings.
\end{abstract}


\section{Introduction}
Federated Learning (FL) is a Machine Learning (ML) paradigm that benefits from collaborative learning while preserving data privacy~\cite{mcmahan2023communicationefficientlearningdeepnetworks}. In the traditional FL architecture, a central server is responsible for aggregating the global model, referred to as Centralized FL (CFL). Nevertheless, CFL is susceptible to a single point of failure issue. Moreover, the bottleneck of the aggregation server can impede the overall model training and aggregation process~\cite{Mart_nez_Beltr_n_2023}. In response to these limitations, Decentralized FL (DFL) has been introduced as an alternative that diminishes the distinction between servers and clients. In DFL, nodes share their local models with neighboring nodes, enabling them to aggregate models based on both their own local models and those of their neighbors, thereby facilitating collaborative learning and privacy preservation~\cite{Mart_nez_Beltr_n_2024}.

However, the data distribution at each node strongly affects the performance of both CFL and DFL models \cite{MA2022244}. When the data does not meet the Independent and Identically Distributed (IID) condition at each node, FL may encounter the client drift issue. This occurs when the optimization objective of the local model differs from that of the entire federation, preventing the aggregated model from converging to the global optimum \cite{jiang2023improving}. Additionally, non-IID data raises security concerns, as popular robust aggregation techniques such as Krum~\cite{blanchard2017machine} and Sentinel~\cite{feng2024sentinel} rely on the distance or similarity between local models. In non-IID scenarios, the differences between local models increase. Thus, it is challenging for aggregation functions to distinguish whether the increased differences are from malicious nodes or non-IID~\cite{feng2024leveraging}.

Research has been dedicated to addressing client drift issues in non-IID settings~\cite{karimireddy2021scaffold}. These works have followed two divergent paths. The first seeks to alleviate client drift by optimizing the global model to achieve better alignment with the global optimum~\cite{jeong2023communicationefficient}. Conversely, the second capitalizes on client drift, positing that local models can be improved by integrating knowledge from other participating clients \cite{wu2024fedbiotllmlocalfinetuning}. However, most existing approaches assume the presence of a central entity with a comprehensive overview of the federation, facilitating the aggregation of local models and guiding optimization decisions. However, this assumption is incompatible with DFL, where no single node possesses a global perspective. Instead, nodes rely solely on information exchanged with their immediate neighbors. This decentralized nature amplifies the difficulty of effectively addressing the non-IID challenge in DFL.

To cope with the challenges presented by non-IID data in DFL, this paper proposes a novel aggregation algorithm called Federated Entropy Pooling (FedEP) inspired by the entropy pooling algorithm in the finance sector. The proposed method comprises three main phases. Firstly, nodes fit their local data distribution using Gaussian Mixture Models (GMM) before the training process begins, sharing statistical characteristics with neighboring nodes. This strategy ensures that only the synthetic statistical representations are communicated, effectively preserving data privacy by preventing the exchange of raw data. In the second phase, nodes endeavor to approximate the global data distribution by integrating the statistical characteristics provided by neighboring nodes. This collaborative approach enables nodes to gain insights into the overall global data distribution within the federation, thereby mitigating the constraints associated with the lack of a centralized aggregation server. Finally, the nodes assess the contribution of each local model to the global model by employing Kullback-Leibler (KL) divergence, which facilitates the determination of the weight of each model during the aggregation process. 

Extensive experiments conducted on various datasets, including MNIST, FMNIST, and CIFAR10, have shown that FedEP significantly accelerates the convergence of DFL models and enhances their performance in various non-IID scenarios. The results provide valuable insights into the strategies for optimizing DFL training from heterogeneous data.

\section{Related Work}
\label{sec:related}
Current research has explored different strategies to reduce the impact of data heterogeneity in FL. These approaches can be divided into two primary categories: enhancing the generalization of the overall model or enhancing the adaptation of individual client models. \tablename~\ref{tab:related} provides an overview of the research aimed at addressing the non-IID challenges in FL.

\subsection{Improving Global Model} 
The main objective of the global model-based approach is to promote the generalization of the aggregated global model across clients' test sets in non-IID scenarios. 


\cite{https://doi.org/10.48550/arxiv.1806.00582} proposed an approach to improve FedAvg for non-IID data by sharing a small portion of local data among all clients before training. However, this method conflicted with the fundamental principle of FL, which aims to keep raw data confidential. Data augmentation techniques used generative models, such as Generative Adversarial Networks (GAN), to locally create additional data samples. In this direction, \cite{jeong2023communicationefficient} presented FAug, where clients uploaded a small amount of data to a server, which then generated a synthetic IID dataset using GAN and sent it back to the clients. However, this approach still required some raw data to be uploaded to the server, raising privacy concerns. \cite{Wang2020OptimizingFL} proposed a client selection strategy that selected clients who participated in the training process with reinforcement learning in each round to mitigate the bias introduced by non-IID data and accelerate convergence.

Knowledge Distillation (KD) was also commonly applied to address non-IID data challenges. \cite{lin2021ensemble} introduced FedDF, an ensemble distillation approach for model aggregation using unlabeled or synthetic data generated by GAN \cite{NIPS2014_5ca3e9b1}. Similarly, \cite{Itahara_2023} proposed a Distillation-Based Semi-Supervised Federated Learning (DS-FL) algorithm that updated parameters through KD using half-private labeled and half-shared unlabeled datasets. Data representation techniques could also be used to mitigate non-IID challenges. \cite{scott2024improvedmodellingfederateddatasets} utilized Mixtures-of-Dirichlet-Multinomials to generate synthetic client data representation and fit a data distribution that mirrored the federation.

\begin{table}[t]\centering
\caption{Existing Methods in FL for Mitigating Data Heterogeneity}
\label{tab:related}
\resizebox{\columnwidth}{!}{%
\newcolumntype{L}[1]{>{\raggedright\let\newline\\\arraybackslash\hspace{0pt}}m{#1}}
\setlength\tabcolsep{1.5pt}
\begin{tabular}{L{1in}L{0.8in}L{0.8in}L{0.8in}c}
\toprule
\textbf{Category} &\textbf{Type} &\textbf{Method} &\textbf{Technique} &\textbf{Schema}  \\ 
\hline
& Data Sharing & ~\cite{https://doi.org/10.48550/arxiv.1806.00582} & Share Raw Data & CFL \\ \cmidrule{2-5}
& Data Augmentation & FAug~\cite{jeong2023communicationefficient}  & GAN & CFL \\\cmidrule{2-5}
Improving Global Model & Client Selection & \cite{Wang2020OptimizingFL} & Reinforcement Learning & CFL \\ \cmidrule{2-5}
& KD & FedDF~\cite{lin2021ensemble} & Federated Distillation & CFL \\
& & DS-FL~\cite{Itahara_2023} & Semi-Supervised Learning & CFL \\ \cmidrule{2-5}
& Data Representation & \cite{scott2024improvedmodellingfederateddatasets}  &  Mixtures-of-Dirichlet-Multinomials & CFL \\

\cmidrule{1-5}
& Model Adaptation  & ADAGRAD, ADAM, YOGI~\cite{reddi2020adaptive} & Adaptive Optimizers & CFL \\\cmidrule{2-5}

Adapting Local Model& Regularization  & SCAFFOLD~\cite{karimireddy2021scaffold} & Client-variance Reduction & CFL \\
&   & FedProx~\cite{li2020federated} & Proximal Term & CFL \\\cmidrule{2-5}
& Personalization  & FedBiOT~\cite{wu2024fedbiotllmlocalfinetuning} & Local Fine-tuning & CFL \\
\\\cmidrule{1-5}

 This work & & FedEP & Entropy Pooling & DFL \\
    \bottomrule
    \end{tabular} 
}
\end{table}

\subsection{Adapting Local Model}
Global model optimization algorithms typically necessitate the exchange of raw data or the retention of a subset of data on the server side, which poses significant privacy concerns. Thus, approaches to optimizing local model training have been proposed to address the non-IID issue while preserving client-side data privacy.

\cite{reddi2020adaptive} proposed adaptive optimizers, such as ADAGRAD, ADAM, and YOGI, which dynamically adjusted learning processes based on clients' data characteristics, improving convergence by fine-tuning learning rates and regularization parameters. \cite{karimireddy2021scaffold} introduced Stochastic Controlled Averaging (SCAFFOLD), an algorithm that reduced client drift in local updates through a client-variance reduction technique, leveraging the similarity among client data to accelerate convergence. FedProx~\cite{li2020federated} extended FedAvg by incorporating a proximal term into the objective function to reduce discrepancies between local and global models. Similarly, \cite{wu2024fedbiotllmlocalfinetuning} proposed FedBiOT, a local fine-tuning method that adapted global parameters received from the server to local data characteristics, enhancing performance while preserving data privacy.

Despite these advancements, the analysis summarized in \tablename~\ref{tab:related} underscores a significant research gap. Most non-IID optimization studies have focused on CFL. While DFL has garnered growing interest, the challenge of data heterogeneity in this decentralized context remains unresolved.

\section{FedEP Algorithm}
\label{sec:fedep}
To address the challenges posed by non-IID data in DFL, this paper introduces a model aggregation function termed FedEP. This section starts from the problem statement and subsequently describes the FedEP algorithm, detailing how it works and the phrases involved.

\subsection{Problem Statement}
In a DFL system involving $N$ participants $\mathbb{C}$, depending on the overlay topology, each participant (denoted $c_j$) is only connected to a specific subset of $K$ participants, known as neighbors, represented as $\mathbb{C}_j$ $\subset$ $\mathbb{C}$. Consequently, each participant aggregates its local model ($w$) with the models shared by its neighbors' models. The objective of DFL local model training is to minimize the test loss ($F(w)$), as defined in Equation{~\eqref{eq:dfl_goal}}.
\setlength\itemsep{0em}
\begin{equation}
    \underset{w}{\text{min }} { F(w) = \sum_{k=1}^{|K|} \alpha_k F_k(w)}
    \label{eq:dfl_goal}
\end{equation}

where $\alpha_k$ is the wights of each client $k$ to the aggregated model, and $\alpha_k \geq 0 $ and $\sum_{k=1}^{|K|}\alpha_k$ = 1.

As shown in Equation{~\eqref{eq:dfl_goal}}, determining the parameter $\alpha$ plays a crucial role in model aggregation. The conventional FedAvg algorithm uses arithmetic averaging, assigning equal weights to all participating models. However, in non-IID settings, models trained on local datasets often exhibit significant variability. As a result, simple arithmetic averaging can cause the aggregated model to deviate from the global optimum. To address this challenge, this paper explores the design of an optimal weight assignment algorithm that improves aggregation by approximating the global optimum while dynamically adjusting weights based on the distribution of local datasets.

\begin{figure}[t]
    \centering
    \includegraphics[width=1\columnwidth]{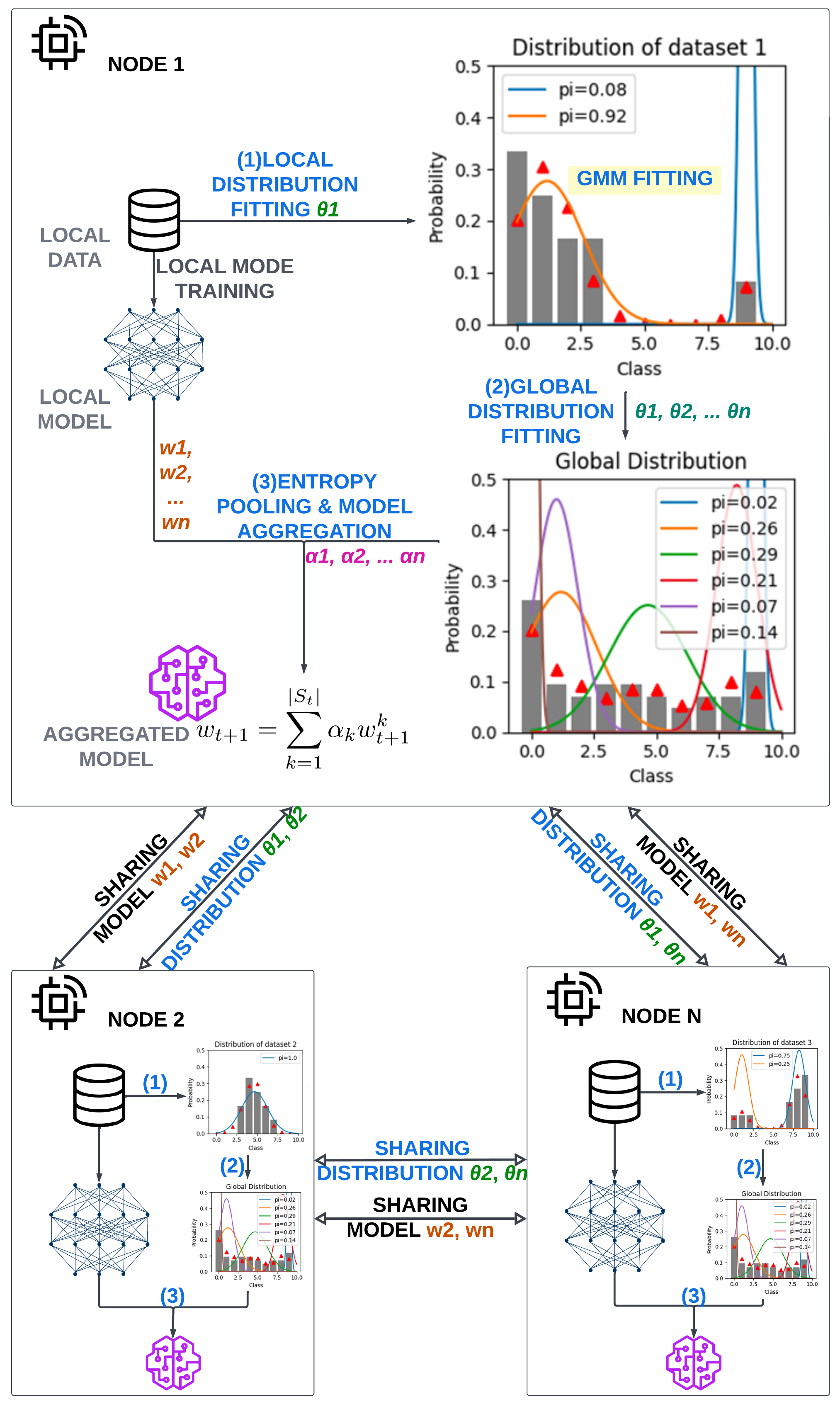}
    \caption{Overview of the FedEP Process, the \textcolor{RoyalBlue}{BLUE} phases are newly introduced by FedEP.}
    \label{fig:arch}
\end{figure}

\subsection{Algorithm Description}
Entropy pooling is a novel approach in the finance sector designed to aggregate diverse opinions on the market~\cite{meucci2010fully}. Consider a scenario with $n$ investors, each with unique opinions ($o_i \in O$) shaped by their specific areas of interest and valid market observations. These views are inherently heterogeneous due to the varied focus and expertise of the investors. To address the challenge of aggregating such diverse opinions, the entropy pooling algorithm calculates relative entropy to construct a global market observation. Based on this global observation, weights are assigned to each investor’s viewpoint, resulting in an aggregated set of investment opinions. This process is similar to the non-IID challenge in DFL, where individual nodes encounter heterogeneous data, develop independent understandings, and lack prior knowledge of global data distribution. Inspired by this analogy, this paper applies the entropy pooling technique to reconstruct the global data distribution using local observations from individual nodes and aggregates their models for improved performance.

The proposed FedEP aggregation algorithm enhances model performance in non-IID scenarios by allowing each node to gain knowledge about the overall federated data distribution and the contribution of individual nodes to the global data, without compromising data privacy. Achieving this requires addressing three fundamental problems: (1) sharing local data distribution information among nodes in a privacy-preserving manner, (2) utilizing shared local distribution information to infer the global data distribution, and (3) assigning aggregation weights to nodes based on both local and global distribution information.

To tackle these issues, the FedEP algorithm employs a three-stage process: (1) pre-training distribution fitting to capture local data characteristics, (2) global data distribution fitting to estimate the federated distribution, and (3) entropy pooling and weighted aggregation to optimize the final model performance. 

An overview of the FedEP process is illustrated in \figurename~\ref{fig:arch}. The phrases highlighted in blue represent new elements introduced by the FedEP algorithm, in addition to the standard DFL processes. A detailed explanation of these phrases is provided in the subsequent sections.

        



\subsubsection{\textbf{Phase 1: Pre-train Distribution Fitting.}}
The primary objective of the first phase is to develop an effective and privacy-preserving mechanism for disseminating the local data distribution information of nodes to other nodes within the DFL network. To achieve this, FedEP employs an innovative approach integrating a GMM with the Expectation-Maximization (EM) algorithm to fit the local data distribution $P_k$. To preserve data privacy, nodes are instructed to share only the parameters $\bm{\theta}_k$ of their respective fitted local distributions.

Each client fits a GMM with an EM method on its local distribution $P_k$ and then sends the resulting GMM parameters $\bm{\theta}_k$ and its number of data samples $N_k$ to the aggregator(s). Algorithm~{\ref{alg:gmm}} shows the main steps in this local distribution fitting phase.

\begin{algorithm}
\caption{\textit{Pre-train Distribution Fitting Algorithm}.}
\begin{algorithmic}[1]
\Require \( \rho \in (0,1]\) is the maximum component fraction; \( M_{\text{max}} \) is the upper bound for the number of mixture models;$M$ is the total number of Gaussian models used in the GMM.
\Function{Pre-trainDistributionFitting}{$\bm{y_k},\rho$}\label{func:pdf}
        \State Initialize $M_{\text{max}} = \lceil \rho \times |\mathcal{Y}_k|\rceil$
        \State Initialize $\hat{M}_k$ and $\hat{\bm{\theta}}_k$
        \For{$M$ in $[1, M_{\text{max}}]$}
            \State $L(\bm{\theta}_M),\bm{\theta}_M \gets \Call{ExpectationMax}{M, \bm{y_k}}$
            \State $BIC_M = -2 \ln(L(\bm{\theta}_M)) + |M| \times \ln(N)$
            \If{BIC of the current model is lower than previous models}
                \State Update $\hat{M}_k = M$ and $\hat{\bm{\theta}}_k = \bm{\theta}_M$
            \EndIf
        \State \Return $\hat{\bm{\theta}}_k$
    \EndFor
\EndFunction

\vspace{1mm}

\Function{ExpectationMax}{$M,\bm{y_k}$}\label{func:em}
    \State Initialize $\bm{\theta}^0 = \{\pi^0, \mu^0,\sigma^{2(0)}\}$ 
    \State E-step: with $\bm{\theta}^t$, get the latent variable $\hat{\gamma}_{j,m}$ 
    \State \hspace{\algorithmicindent} and $Q(\bm{\theta}^t)$, s.t. $\sum_{n=1}^M \pi_m = 1$
    \State M-step: $\bm{\theta}_{i+1} \gets \arg\max{\bm{\theta}(Q + \lambda(\sum_{n=1}^M \pi - 1))}$
    \State Repeat E, M until convergence
    \State \Return $L(\bm{\theta}), \bm{\theta}$
\EndFunction
\end{algorithmic}
\label{alg:gmm}
\end{algorithm}

A GMM is a parametric probability density function represented as a weighted sum of Gaussian component densities, as shown in Equation~\ref{eq:gmm}:

\begin{equation}
P(\mathbf{y}_k \mid \bm{\theta}_k) = \sum_{m=1}^{M} \pi_m \mathcal{N}(\mu_m, \sigma^2_m), 
\label{eq:gmm}
\end{equation}
where \(\pi_m > 0\) for all \(m = 1, \dots, M\), and \(\sum_{m=1}^M \pi_m = 1\). \( \mathbf{y}_k \) is the label vector \( [y_1, y_2, \ldots, y_{N_k}] \) of client \( k \). \( P(\mathbf{y}_k|\bm{\theta}) \) is the probabilistic distribution of the labels of client \( k \) given \( \mathbf{\bm{\theta}} \). \( \mathbf{\bm{\theta}} \) is a \( 3 \times M \) coefficients matrix \( [\bm{\pi}, \bm{\mu}, \bm{\sigma}^2] \). The hyper-parameter \( M \) is the total number of used Gaussian Models in the GMM. The mixture coefficient vector \( \bm{\pi} = [\pi_1, \pi_2, \ldots, \pi_M] \), with each element as the coefficient of the \( m \)-th Gaussian distribution. The vector \( \bm{\mu} = [\mu_1, \mu_2, \ldots, \mu_M] \), with each element as the mean of the \( m \)-th Gaussian distribution. The vector \( \bm{\sigma}^2 = [\sigma^2_1, \sigma^2_2, \ldots, \sigma^2_M] \), with each element as the variance of the \( m \)-th Gaussian distribution. Note that if some observations \( \bm{y} \sim \mathcal{N}(\mu_m, \sigma^2_m) \), their PDF is as Equation~\ref{eq:gmm_pdf}:

\begin{equation}
f(\bm{y}|\mu_m,\sigma^2_m) = \frac{1}{\sigma_m\sqrt{2\pi}} e^{-\frac{1}{2}\left(\frac{y-\mu_m}{\sigma_m}\right)^2}
\label{eq:gmm_pdf}
\end{equation}

Identifying the optimal parameters of a GMM presents a significant challenge, particularly in determining the appropriate number of Gaussian distributions that comprise the original dataset. This work employs the EM algorithm to find the optimal parameters of the GMM by incorporating a latent variable matrix $\bm{\gamma}$, thereby enabling a more accurate representation of the local dataset's distribution. The latent variable matrix $\bm{\gamma}$ is defined in Equation \ref{eq:varibalematrix}:

\begin{equation}
\bm{\gamma}_{n,m} =
\begin{pmatrix}
    \gamma_{11} & \gamma_{12} & \cdots  & \gamma_{1M} \\
    \gamma_{21} & \gamma_{22} & \cdots & \gamma_{2M} \\
    \vdots & \vdots & \ddots & \vdots   \\
    \gamma_{N1} & \gamma_{N2} &  \cdots & \gamma_{NM}
\end{pmatrix}_{N \times M}
\label{eq:varibalematrix}
\end{equation}
Each elements $\gamma_{n,m}$ represents whether the $n$-th data sample belongs to the $m$-th Gaussian Model, as shown in Equation~\ref{eq:gamma}:

\begin{equation}
\centering
\begin{split}
   \gamma_{n,m} &:=
\begin{cases} 
    1 & \text{if } y_n \sim \mathcal{N}(\mu_m, \sigma^2_m), \\
    0 & \text{else}.
\end{cases}  \\
\quad \sum^M_{m=1} \gamma_{nm} &=1, m\in[1,M], n\in[1,N].
\end{split}
\label{eq:gamma}
\end{equation}

The EM algorithm consists of four steps:

\textit{\textbf{Step 1 (Initialization)}}: Initialize the coefficients matrix $\bm{\theta}^{(0)} = [\bm{\pi^{(0)}},\bm{\mu^{(0)}},\bm{\sigma^{2(0)}}]$.

\textit{\textbf{Step 2 (E-step)}}: Based on $\bm{\theta}^{(t)}$, firstly calculate the estimated latent variables matrix $\hat{\bm{\gamma}}^{(t)}$:
\begin{equation}
\hat{\bm{\gamma}}^{(t)}:=  
\begin{pmatrix}
    \hat{\gamma}_{11} & \hat{\gamma}_{12}  & \cdots & \hat{\gamma}_{1M} \\
    \hat{\gamma}_{21} & \hat{\gamma}_{22} & \cdots & \hat{\gamma}_{2M} \\    
    \vdots & \vdots & \ddots &  \vdots \\
    \hat{\gamma}_{N1} & \hat{\gamma}_{N2} & \cdots & \hat{\gamma}_{NM}
\end{pmatrix}_{N \times M}
\end{equation}
where
\begin{equation}
\begin{split}
   \hat{\gamma}_{nm} &:= E(\gamma_{n,m}|y_n,\bm{\theta}^{(t)}) \\ 
   &= \frac{\pi_m \mathcal{N}(y_n|\mu^{(t)}_m,(\sigma^{2})^{(t)}_m)}{\sum_{m=1}^M \pi_m \mathcal{N}(y_n|\mu^{(t)}_m,(\sigma^{2})^{(t)}_m)}
\end{split}
\end{equation}

Based on $\bm{\theta}^{(t)}$, the log-likelihood function $Q(\bm{\theta}^{(t)})$ is:

\begin{equation}
\centering
\begin{split}
Q(\bm{\theta}^{(t)})
&:=ln(L(\bm{\theta}^{(t)})) \\
&= \sum^M_{m=1}\{(\sum^{N}_{n=1}\gamma_{nm}) ln(\pi_m)+ \\ 
   &\quad \sum^{N}_{n=1}\gamma_{nm}ln(\mathcal{N}(y_n|\mu^{(t)}_m,(\sigma^{2})^{(t)}_m))\}, \\
   \quad s.t.\sum^M_{m=1} \pi_m &= 1
\end{split}
\end{equation}

\textit{\textbf{Step 3 (M-step)}}: To maximize the log-likelihood function $Q(\bm{\theta}^{(t)})$ found in the E-step, it is necessary to calculate the derivatives of $Q(\bm{\theta}^{(t)})$ with respect to $\bm{\pi},\bm{\mu},\bm{\sigma^2}$

\begin{equation}
\centering
\begin{split}
\hat{\bm{\pi}}_m = \frac{\sum_{n=1}^{N} \hat{\gamma}_{nm}}{\sum_{m=1}^{M} \sum_{n=1}^{N} \hat{\gamma}_{nm}}, 
\quad \hat{\bm{\mu}}_m = \frac{\sum_{n=1}^{N} \hat{\gamma}_{nm} \bm{y}_n}{\sum_{n=1}^{N} \hat{\gamma}_{nm}}, \\
\quad\hat{\bm{\sigma}}^2_m = \frac{\sum_{n=1}^{N} \hat{\gamma}_{nm} (\bm{y}_n - \hat{\bm{\mu}}_m)(\bm{y}_n - \hat{\bm{\mu}}_m)^T}{\sum_{n=1}^{N} \hat{\gamma}_{nm}}
\end{split}
\end{equation}

\textit{\textbf{Step 4 (Iterate until Convergence)}}: Repeat Steps 2 and 3 until convergence: the change of the value of $Q(\bm{\theta}^{(t)})$ between iterations is below a predetermined threshold. Export $\bm{\theta}^* = [\hat{\bm{\pi}}_m,\hat{\bm{\mu}}_m,\hat{\bm{\sigma}}^{2}_m]$ as the result.

To determine the optimal number of components for each distribution $P_k$, an iterative process over a range from 1 to $M_{max}$ is conducted, which selects the model with the lowest Bayesian Information Criterion (BIC) value as the ideal $\hat{M}$. A hyper-parameter $\rho$, which represents the maximum component fraction, is utilized to determine the upper limit $M_{\text{max}}$ for the number of mixture models. Specifically, $M_{\text{max}}$ is defined as $\rho$ times the total number of distinct label classes $\mathcal{Y}_k$, where $\rho$ ranges from 0 to 1 and is typically set to 0.5 as a default value.

\begin{equation}
\centering
\begin{split}
\hat{M} = \underset{ M\in\mathbb{Z}}{\arg\min} \{-2 \ln(\mathcal{L}(\bm{\theta}^*)) + |M| \times \ln(N)\}, \\
\quad 1<M<  \lceil \rho\times |\mathcal{Y}_k|\rceil,
\quad \rho \in (0,1]
\end{split}
\end{equation}

The resulting $\bm{\theta}^*$ serves as an effective summary of the heterogeneous data distributions, enabling efficient estimation of the global distribution. Unlike directly sharing raw data or full data distributions, transmitting the parameters of a fitted distribution significantly reduces communication overhead while preserving data privacy.

\subsubsection{\textbf{Phase 2: Global Distribution Estimation.}} Upon receiving \( K \) fitting results \( \bm{\theta}_k^* \), the node (acting as both trainer and aggregator) estimates the global model distribution as:

\begin{equation}
\centering
\begin{split}
P(Y|\bm{\theta}_1^*, \ldots, \bm{\theta}_K^*, N_1, \ldots, N_K) = \\ \quad \sum_{k=1}^K p_k \left[ \sum_{m=1}^M \pi_{mk} \mathcal{N}(\mu_{mk}, \sigma^2_{mk}) \right]
\end{split}
\end{equation}
where \( p_k = \frac{N_k}{\sum_{k=1}^{K} N_k} \). The global distribution estimation is the weighted average of $K$ GMMs, and $\sum_{k=1}^K p_k \sum_{m=1}^M \pi_{mk} = 1$.

Phase 2 addresses the problem of aggregating what each node has seen in its local data to understand the broader picture. By combining observations from neighbors, each node obtains an estimate of the overall data distribution. The subsequent phase focuses on optimizing the aggregation of local models by leveraging the estimated global and local data distributions.

\subsubsection{\textbf{Phase 3: Entropy Pooling and Weighted Aggregation.}} 
KL-divergence is used to quantify the difference between two probability distributions, as defined in Equation~\ref{eq:kld}.
\begin{equation}
   \text{KLD}_k(P \Arrowvert P_k) = \sum_{y \in \mathcal{Y}} P(y) \left(\ln(P(y)) - \ln(P_k(y))\right) 
\label{eq:kld}
\end{equation}

When considering \( P_k \), the distribution of individual client \( k \), as an approximation of the overall distribution \( P \), FedEP calculates the pooled KL-divergence for each client \( k \) by obtaining \( K \) KL-divergences and determining the attention coefficient:
\begin{equation}
\begin{split}
\alpha_k &= \frac{\text{KLD}_k}{\sum_{k=1}^K \text{KLD}_k} \\
\quad &= \frac{\sum_{y \in \mathcal{Y}} P(y) \left(\ln(P(y)) - \ln(P_k(y))\right)}{\sum_{k=1}^K \sum_{y \in \mathcal{Y}} P(y) \left(\ln(P(y)) - \ln(P_k(y))\right)}
\end{split}
\end{equation}

The clients send their updated weights back to the Aggregator(s), which then aggregate(s) these weights into $w_{t+1}$ using a weighted average with weights being the pooled KL-divergence $\alpha_k$:
\begin{equation}
w_{t+1} = \sum_{k=1}^{|S_t|} \alpha_{k} w_{t+1}^k
\end{equation}

Through the three phrases, each node utilizes its own data distribution information and the data distribution from neighbors to infer the global data distribution. Using the entropy pooling technique, weights for the participating models are calculated, enabling the aggregation process to optimize model performance in non-IID scenarios in a privacy-preserving manner.

\section{Deployment and Experiments}
\label{sec:experiments}

To evaluate the performance of the proposed FedEP algorithm in non-IID environments, experiments are conducted comparing it against state-of-the-art algorithms across various datasets with differing levels of non-IID heterogeneity.

\textit{Fedstellar} \cite{Mart_nez_Beltr_n_2024} was employed as the platform for executing DFL scenarios. Implemented in Python, Fedstellar offers Docker-based containerized simulations for FL. The platform supports the creation of federations by allowing customization of parameters such as the number and type of devices training FL models, the datasets each participant utilized, and deep learning algorithms employed.

To model non-IID data, Dirichlet distribution was utilized to generate data distribution for each node. The probability density function of the Dirichlet distribution is given by \eqref{eq:pdf}:

\begin{equation}
f(\boldsymbol{p}; \boldsymbol{\alpha}) = \frac{1}{B(\boldsymbol{\alpha})} \prod_{i=1}^K p_i^{\alpha_i - 1}
,
B(\boldsymbol{\alpha}) = \frac{\prod_{i=1}^K \Gamma(\alpha_i)}{\Gamma\left(\sum_{i=1}^K \alpha_i\right)}
\label{eq:pdf}
\end{equation}
where \( \mathbf{p} = (p_1, p_2, \ldots, p_k) \) is a point in the probability simplex (\( p_i \geq 0 \) and \( \sum_{i=1}^{k} p_i = 1 \)).

The variance of $p_i$ is dependent on the concentration parameter $\alpha$:
\begin{equation}
\text{Var}(p_i) = \frac{\alpha_i (\sum_{j=1}^{k} \alpha_j - \alpha_i)}{(\sum_{j=1}^{k} \alpha_j)^2(\sum_{j=1}^{k} \alpha_j + 1) }
\end{equation}

As the $\alpha$ increases, the term $\alpha_i (\sum_{j=1}^{k} \alpha_j - \alpha_i)$ will increase at a slower rate compared to the denominator, leading to a decrease of $\text{Var}(p_i)$ and $p_i$ closer to $\mathbf{E}[p_i]$. The Dirichlet distribution is adept at modeling data heterogeneity through varying $\alpha$ values. 

\subsection{Experiments Configuration}

In this work, the FedEP was implemented in Python. A set of experiments was executed with the following setups:
\begin{itemize}
    \item The federation consists of 10 fully connected nodes, with 10 rounds of training (each round with 3 epochs).  
    \item MNIST~\cite{lecun2010mnist}, Fashion-MNIST~\cite{xiao2017fashion}, and CIFAR10 \cite{krizhevsky2009learning} are used for experiments. 
    \item The chosen model topology for MNIST and Fashion-MNIST datasets is a Multi-Layer Perceptron (MLP) containing two fully connected layers with 256 and 128 neurons. For CIFAR10, SimpleMobileNet \cite{sinha2019thin} is used. 
    \item FedAvg, SCAFFOLD, FedProx, and FedEP (this work) are implemented, evaluated, and compared.
    \item Averaged test dataset F1-Score across all nodes is used as a metric to evaluate the performance.
\end{itemize}

This paper introduces two types of non-IID scenarios to provide a more comprehensive evaluation of the algorithms in non-IID environments. The first is \textbf{pure non-IID}, where all nodes adopt the same Dirichlet $\alpha$ parameter for local data sampling. The second is \textbf{mixed non-IID}, where nodes in the federation use different $\alpha$ values for local data sampling.

\subsection{Pure Non-IID Scenario}
For the Pure non-IID scenario, the training datasets are generated using Dirichlet $\alpha$ values of 20, 1, and 0.1.

The experimental results are detailed in \tablename~\ref{tab:pure-f1}. It is evident that at a lower degree of non-IID (alpha=20), the selected algorithms exhibit comparable performance in terms of F1-Score across the datasets.

\begin{table}[t]
\centering
\caption{F1-Score for MNIST, Fashion-MNIST, CIFAR10 in Pure Non-IID Scenarios}
\label{tab:pure-f1}
\resizebox{\columnwidth}{!}{%
\begin{tabular}{lllll}
\toprule
\textbf{Dataset} & \textbf{Aggregation} & $\alpha=20$ &  $\alpha=1$ & $\alpha=0.1$ \\
\toprule
& FedAvg & 0.721  $\pm$  0.016 & \textbf{0.693  $\pm$  0.020} & 0.451  $\pm$  0.013\\
CIFAR10 & FedProx & 0.656  $\pm$  0.013 & 0.656  $\pm$  0.012 & 0.495  $\pm$  0.022\\
& \textbf{FedEP} & \textbf{0.743  $\pm$  0.015} & 0.684  $\pm$  0.018 & \textbf{0.583 $\pm$ 0.016}\\
& SCAFFOLD & 0.742  $\pm$  0.022 & 0.689  $\pm$  0.013 & 0.566  $\pm$  0.015\\ \hline
& FedAvg & 0.899  $\pm$  0.013 & 0.891  $\pm$  0.007 & 0.855  $\pm$  0.014\\
Fashion- & FedProx & 0.815  $\pm$  0.009 & 0.813  $\pm$  0.012 & 0.764  $\pm$  0.017\\
MNIST & \textbf{FedEP} & \textbf{0.903  $\pm$  0.016} & \textbf{0.895  $\pm$  0.011} & \textbf{0.859  $\pm$  0.017}\\
& SCAFFOLD & 0.903  $\pm$  0.014 & 0.893  $\pm$  0.010 & 0.853 $\pm$  0.013\\ \hline
& FedAvg & 0.965  $\pm$  0.008 & 0.958  $\pm$  0.007 & 0.891  $\pm$  0.013\\
MNIST & FedProx & 0.902  $\pm$  0.006 & 0.899  $\pm$  0.010 & 0.843  $\pm$  0.014\\
& \textbf{FedEP} & 0.966  $\pm$  0.008 & \textbf{0.964  $\pm$  0.007} & \textbf{0.924  $\pm$  0.007}\\
& SCAFFOLD & \textbf{0.969  $\pm$  0.007} & 0.959  $\pm$  0.009 & 0.915  $\pm$  0.011\\ \toprule
\end{tabular}%
}
\end{table}

\begin{figure}[t]
    \centering
    \includegraphics[width=1\columnwidth]{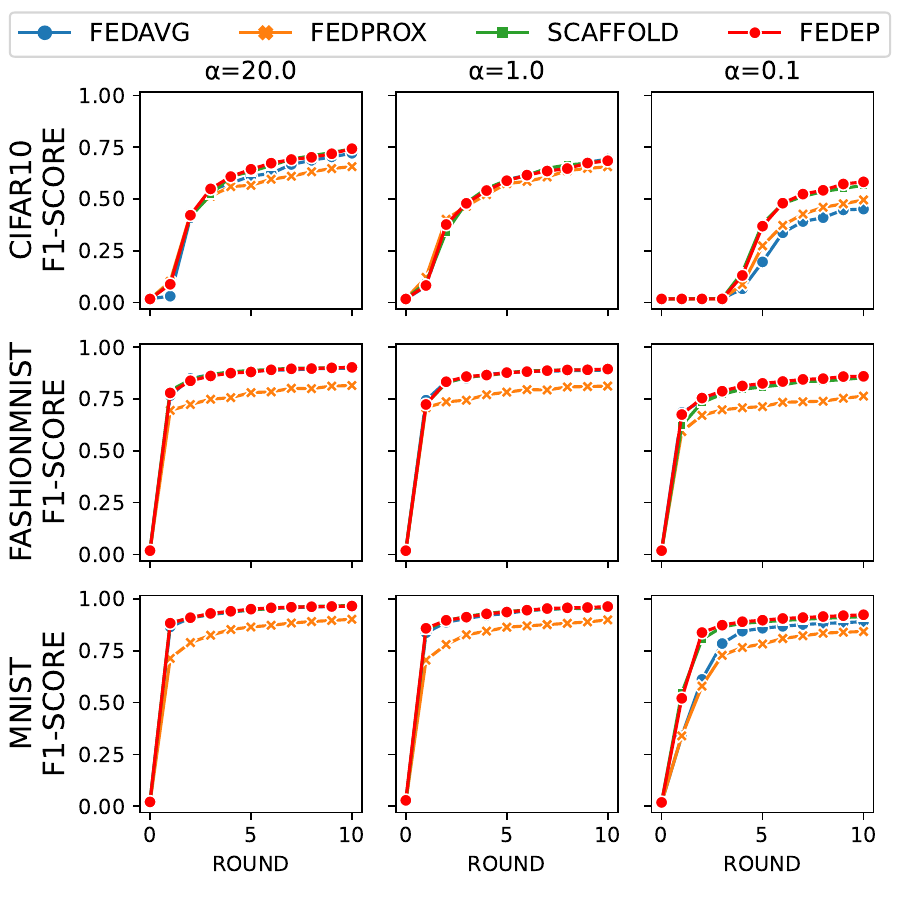}
    \caption{In the pure non-IID scenario, the F1-Score for each round for MNIST, Fashion-MNIST, CIFAR10}
    \label{fig:pure_round}
\end{figure}

Specifically, FedEP surpasses the other algorithms by approximately 0.05 of F1-Score on the MNIST and FashionMNIST datasets when Dirichlet $\alpha$ is reduced to 1. In the extreme scenario where Dirichlet $\alpha$ is equal to 0.1, FedEP's advantage becomes even more evident, demonstrating improvements across all three datasets.

In a more detailed analysis of algorithm performance across varying rounds, as depicted in \figurename~\ref{fig:pure_round}, it is evident that FedEP exhibits superior convergence speed, typically within the first three rounds, accompanied by reduced oscillatory behavior. In contrast, baseline algorithms generally necessitate five rounds to attain convergence and display greater volatility. Thus, in the pure non-IID scenarios, FedEP outperforms its effectiveness by achieving better results in shorter convergence rounds.

\begin{table}[t]
\centering
\caption{F1-Score for MNIST, Fashion-MNIST, CIFAR10 in Mixed Non-IID Scenarios}
\label{tab:mix-f1}
\resizebox{\columnwidth}{!}{%
\begin{tabular}{lllll}
\toprule
\textbf{Dataset} & \textbf{Aggregation} & $\alpha=[50, 20]$ &  $\alpha=[50, 1]$ & $\alpha=[50, 0.1]$ \\
\toprule
  &  FedAvg  &  0.528 $\pm$  0.014  &  0.342 $\pm$  0.014  &  0.026 $\pm$  0.003 \\
CIFAR10  &  FedProx  &  0.581 $\pm$  0.018 &  0.399 $\pm$  0.011  &  \textbf{0.103 $\pm$  0.005} \\
  &  \textbf{FedEP}  &  \textbf{0.662 $\pm$  0.012} &  \textbf{0.471 $\pm$  0.014}  &  0.018 $\pm$  0.001 \\
  &  SCAFFOLD  &  0.621 $\pm$  0.012 &  0.315 $\pm$  0.010  &  0.017 $\pm$  0.001 \\ \hline
  &  FedAvg  &  0.801 $\pm$  0.004 &  0.402 $\pm$  0.012  &  0.017 $\pm$  0.001 \\
Fashion-  &  FedProx  &  0.667 $\pm$  0.019 &  0.275 $\pm$  0.008  &  0.132 $\pm$  0.007 \\
MNIST  &  \textbf{FedEP}  & \textbf{0.833 $\pm$ 0.011}  &  \textbf{0.596 $\pm$  0.017}  &  \textbf{0.211 $\pm$  0.011} \\
  &  SCAFFOLD  &  0.790 $\pm$  0.014 &  0.458 $\pm$  0.011  &  0.017 $\pm$  0.001 \\  \hline
  &  FedAvg  &  0.773 $\pm$  0.015 &  0.319 $\pm$  0.008  &  0.208 $\pm$  0.007 \\
  &  FedProx  &  0.898 $\pm$  0.011 &  0.295 $\pm$  0.009  &  0.017 $\pm$  0.001 \\
MNIST  &  \textbf{FedEP}  &  0.958 $\pm$  0.011 &  \textbf{0.667 $\pm$  0.006}  &  \textbf{0.349 $\pm$  0.015} \\
  &  SCAFFOLD  &  \textbf{0.960 $\pm$  0.008} &  0.511 $\pm$  0.009  &  0.208 $\pm$  0.009 \\

 \toprule
\end{tabular}%
}
\end{table}

\begin{figure}[t]
    \centering
    \includegraphics[width=1\columnwidth]{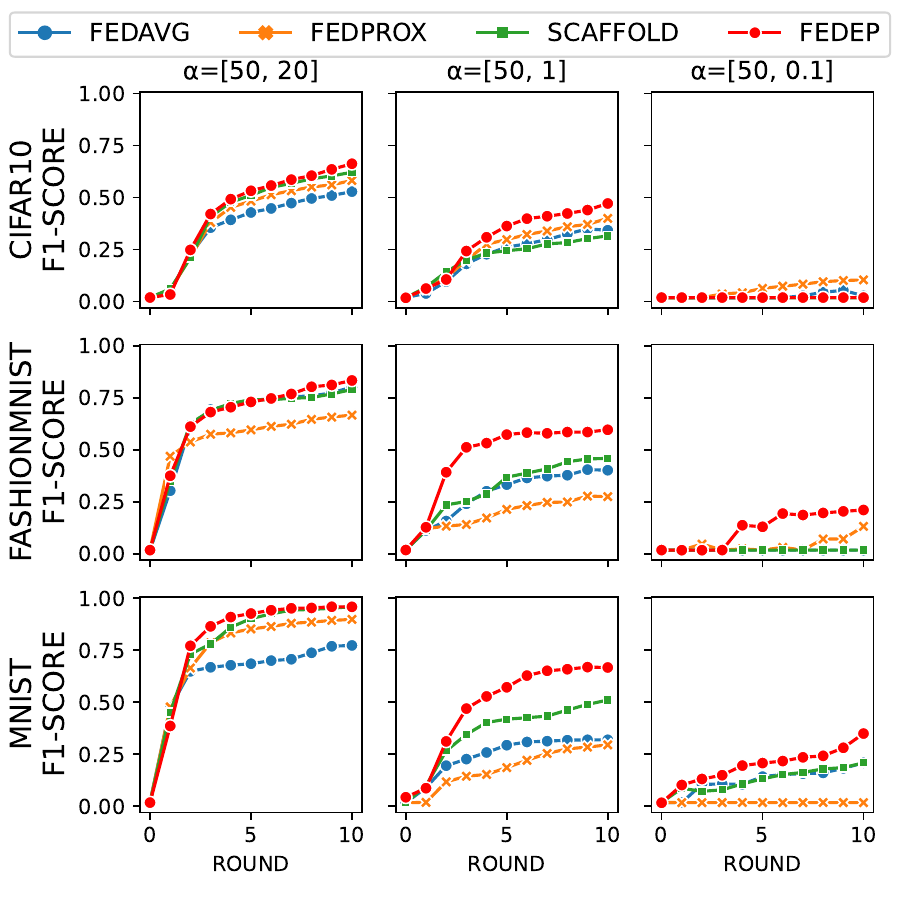}
    \caption{In the mixed non-IID scenario, the F1-Score for each round for MNIST, Fashion-MNIST, CIFAR10}
    \label{fig:mix}
\end{figure}

\subsection{Mixed Non-IID Scenario}
The previous experiment considered all nodes at the same degree of non-IID, i.e., the same Dirichlet $\alpha$ was used for non-IID in one federation. This experiment uses a mixed non-IID scenario, dividing the nodes into two groups and using two different Dirichlet $\alpha$ values for the federation's non-IID modulation. Specifically, 50\% of the nodes with a Dirichlet $\alpha$ value of 50 and the remaining 50\% of the nodes with an $\alpha$  of [20, 1, 0.1] were used to divide the data and discuss the performance of the FedEP versus baseline algorithms in more complex scenarios.

In this experiment, as demonstrated in \tablename~\ref{tab:mix-f1}, FedEP outperforms other aggregation algorithms across the three datasets in various levels of mixed non-IID scenarios. When the Dirichlet $\alpha$ is [50, 20], FedEP achieves an F1-Score advantage of over 0.05 compared to FedAvg on all three datasets. When the Dirichlet $\alpha$ is [50, 1], FedEP demonstrates a significant advantage of over 0.1 compared to other aggregation methods. This indicates that FedEP effectively mitigates the impact of mixed non-IID settings, guiding the models on individual nodes toward the global optimum.

In terms of convergence speed, as shown in \figurename~\ref{fig:mix}, FedEP also has a clear advantage. FedEP's F1-Score converges around the third round of testing, whereas other algorithms require approximately five rounds to reach a similar F1-Score. This advantage becomes even more pronounced as the Dirichlet $\alpha$ decreases.

\section{Discussion}
Experiments indicate the proposed FedEP algorithm can expedite the convergence of the DFL model and improve its performance. This section discusses FedEP's privacy-preserving capabilities, computational and communication overhead, and scalability.

\textbf{\textit{Privacy Perversion}:} The FedEP algorithm preserves privacy in DFL by sharing only synthetic data distribution information, while keeping original data strictly local. Unlike methods that share raw or synthetic data (\textit{i.e.} \cite{https://doi.org/10.48550/arxiv.1806.00582}), FedEP exchanges abstract representations, minimizing privacy risks. While the SCAFFOLD algorithm requires exchanging control variables, FedEP avoids transmitting sensitive gradient-related information, offering comparable privacy safeguards. Thus, FedEP mitigates non-IID challenges while preserving data privacy.

\textbf{\textit{Computational and Communication Overhead}:}
The additional computational complexity of FedEP, compared to the FedAvg algorithm, arises from using the EM algorithm to determine the optimal local model distribution. Assuming each node has a local sample size of \( N \), with \( K \) possible distributions, and each probability density computation has a complexity of \( \mathcal{O}(D) \), the EM algorithm requires \( T \) iterations to converge. Thus, the overall computational complexity is $\mathcal{O}(T \cdot N \cdot K \cdot D)$, which does not significantly increase the computational cost. For communication, only the parameter vectors of the GMM are transmitted, adding a minimal overhead within the byte range. Therefore, FedEP does not significantly increase the communication overhead.

\begin{figure}[t]
    \centering
    \includegraphics[width=1\columnwidth]{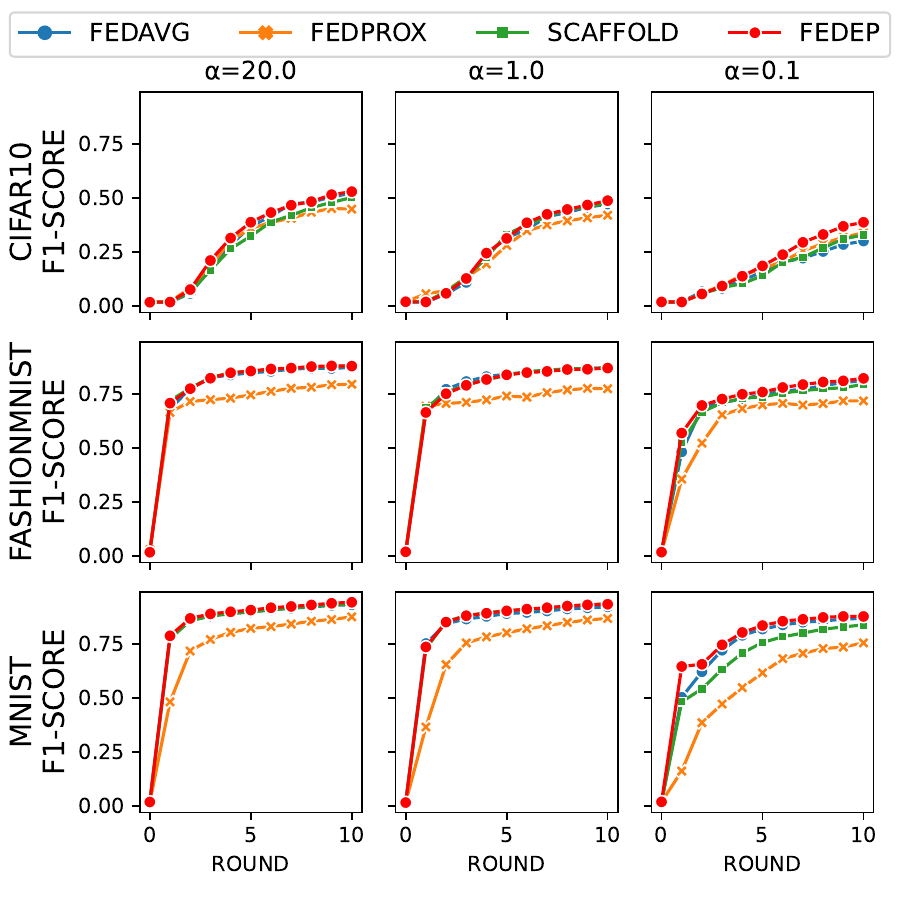}
    \caption{In the 20-nodes pure non-IID scenario, the F1-Score for each round for MNIST, Fashion-MNIST, CIFAR10}
    \label{fig:20nodes}
\end{figure}

\textbf{Scalability:} Experiments were conducted on a federation of 20 fully connected nodes, training on three datasets with Dirichlet $\alpha$ values of 20, 1, and 0.1. As shown in \figurename~\ref{fig:20nodes}, FedEP consistently outperforms reference algorithms in various levels of non-IID settings within a 20-node DFL, demonstrating its superiority. When $\alpha$ equals 20, FedEP achieves F1-Scores similar to FedAvg and SCAFFOLD. As $\alpha$ decreases, FedEP's advantage becomes more pronounced, with an increasing lead over other algorithms. In terms of convergence speed, FedEP achieves optimization targets faster than baseline models. These results indicate that as the number of nodes increases, FedEP maintains its effectiveness, proving its strong scalability and potential for application in large-scale DFL systems.

\section{Conclusion and Future Work}
\label{sec:conclusion}

This work introduces FedEP, an algorithm designed to address the non-IID data issue in DFL in a privacy-preserving manner. By disseminating the data distribution across nodes, FedEP aligns with the overall data distribution within the federation and employs this alignment to determine the weighting of individual nodes during the aggregation process. Comprehensive assessments conducted on the MNIST, Fashion-MNIST, and CIFAR10 datasets under various non-IID conditions underscore the effectiveness of FedEP. Compared with other cutting-edge aggregation algorithms, FedEP consistently exhibits superior performance and faster convergence rates in both pure non-IID and mixed non-IID scenarios. FedEP effectively preserves privacy through its unique mechanism of sharing fitted data distribution. Moreover, its computational and communication overheads are not higher than those of other reference algorithms.

Future research will investigate the impact of malicious attacks, such as poisoning attacks, on the robustness of FedEP. Additionally, there is a growing interest in exploring the application of FedEP in advanced areas such as multi-task learning, where it could potentially enhance performance by effectively handling heterogeneous tasks across distributed nodes.

\balance
\bibliographystyle{IEEEtran} 
\bibliography{ref}

\end{document}